\documentclass{article}

\usepackage{microtype}
\usepackage{graphicx}
\usepackage{subcaption,comment}
\usepackage{booktabs} 

\usepackage{booktabs}   
\usepackage{multirow}   
\usepackage{array}      

\usepackage{tabularx}

\usepackage{hyperref}

\newcommand{\mhead}[1]{{\small  #1}} 

\newcommand{\secref}[1]{Sec.~\ref{#1}}
\newcommand{\figref}[1]{Fig.~\ref{#1}}

\usepackage[accepted]{icml2026}

\makeatletter
\renewcommand{\ICML@appearing}{%
  \textit{Accepted at LM4Plan @ ICML 2026}, Seoul, South Korea, July 11, 2026.%
}
\makeatother

\usepackage{amsmath}
\usepackage{amssymb}
\usepackage{mathtools}
\usepackage{amsthm}

\usepackage{listings}

\usepackage[capitalize,noabbrev]{cleveref}

\theoremstyle{plain}

\theoremstyle{definition}

\theoremstyle{remark}

\usepackage[textsize=tiny]{todonotes}

\icmltitlerunning{VeryTrace: Verifying Reasoning Traces}

\begin{document}

\twocolumn[
  \icmltitle{VeryTrace: Verifying Reasoning Traces through Compilable Formalism and Structured Verification}



  \icmlsetsymbol{equal}{*}

  \begin{icmlauthorlist}
    \icmlauthor{Ninghan Zhong}{yyy}
    \icmlauthor{Ahmet Ege Tanriverdi}{comp}
    \icmlauthor{Kaan Kale}{yyy}
    \icmlauthor{Sriram Vishwanath}{yyy}
    
  \end{icmlauthorlist}

  \icmlaffiliation{yyy}{School of Electrical and Computer Engineering, Georgia Institute of Technology, USA}
  \icmlaffiliation{comp}{Department of Electrical and Computer Engineering, Bogazici University, Turkey}
  

  \icmlcorrespondingauthor{Ninghan Zhong}{nzhong34@gatech.edu}

  \icmlkeywords{Chain of Thought, Reasoning Traces, Domain Specific Languages (DSLs), Auditing Models, Large Language Models}

  \vskip 0.3in
]



\printAffiliationsAndNotice{}  

\begin{abstract}

Multi-step reasoning with Chain-of-Thought (CoT) prompting remains  fragile: logical errors or hallucinations in early steps silently  propagate, producing confident but incorrect conclusions.
This paper presents \textbf{VeryTrace}, a zero-shot verification-and-repair framework that formalizes natural-language reasoning traces into a structured, compilable representation. VeryTrace introduces a Domain-Specific Language (DSL) that: (i) makes 
step dependencies explicit, (ii) mechanizes quantitative content as 
executable expressions, and (iii) structures semantic inferences via 
deduction schemas. Our hybrid 
verifier combines deterministic checks for computational correctness, 
dependency resolution, \& constraint satisfaction with targeted LLM 
audits for non-mechanizable semantic judgments, enabling step-level 
error localization and repair.

Across three diverse domains—competition mathematics (AIME 2025), 
robotics planning (LLM-BabyBench), and kinship reasoning (CLUTRR)—
VeryTrace improves accuracy over zero-shot baselines on state-of-the-art
 LLMs without requiring domain-specific training or in-context examples, 
demonstrating that formalized trace verification achieves both 
precision and generalization.

\end{abstract}

\section{Introduction}
\label{Introduction}

Chain-of-Thought (CoT) prompting has transformed how Large Language Models (LLMs) handle complex reasoning tasks, enabling multi-step derivations that outperform single-shot predictions. However, this capability remains fundamentally brittle. A single arithmetic error, unstated assumption, or logical fallacy at step $k$ can cascade through steps $k+1, k+2, \ldots$, producing a coherent-sounding but incorrect conclusion. This error propagation problem intensifies in domains requiring strict correctness: competition mathematics demands precise symbolic manipulation, planning tasks require invariant satisfaction at every step, and relational reasoning depends on consistent inference chains.

Existing mitigation strategies face a fundamental trade-off: End-to-end verification evaluates only final answers, providing no diagnostic granularity when traces fail. Self-consistency and voting approaches require multiple expensive samples without diagnosing \emph{why} individual traces fail. At the other extreme, formal theorem provers (Lean, Coq, Isabelle) offer rigorous verification but demand domain-specific formalization, expert-level syntax, and extensive manual effort, requirements that do not transfer across problem types. What is missing is a framework that provides step-level verification granularity while maintaining domain-agnostic generalization — one that verifies the \emph{process} of reasoning, not just the outcomes.

We present \textbf{VeryTrace}, a framework that treats reasoning traces as compilable programs that can be verified step-by-step. Drawing inspiration from tactic-style theorem proving (Lean, Isabelle), VeryTrace models reasoning as a sequence of state transitions: each step declares its dependencies, applies a computation or deduction schema, and produces an updated reasoning state containing variable bindings and active constraints. This representation enables two critical capabilities: (i) mechanized verification of the parts of reasoning that \emph{can} be formalized (arithmetic, symbolic evaluation, dependency ordering), and (ii) structured LLM audits for the semantic parts that \emph{cannot} be formalized (natural language constraints, commonsense inferences). Crucially, formalization logic and verification pipeline apply across domains without specialized handlers.

\begin{figure*}[tb]
    \centering
    \includegraphics[width=\linewidth]{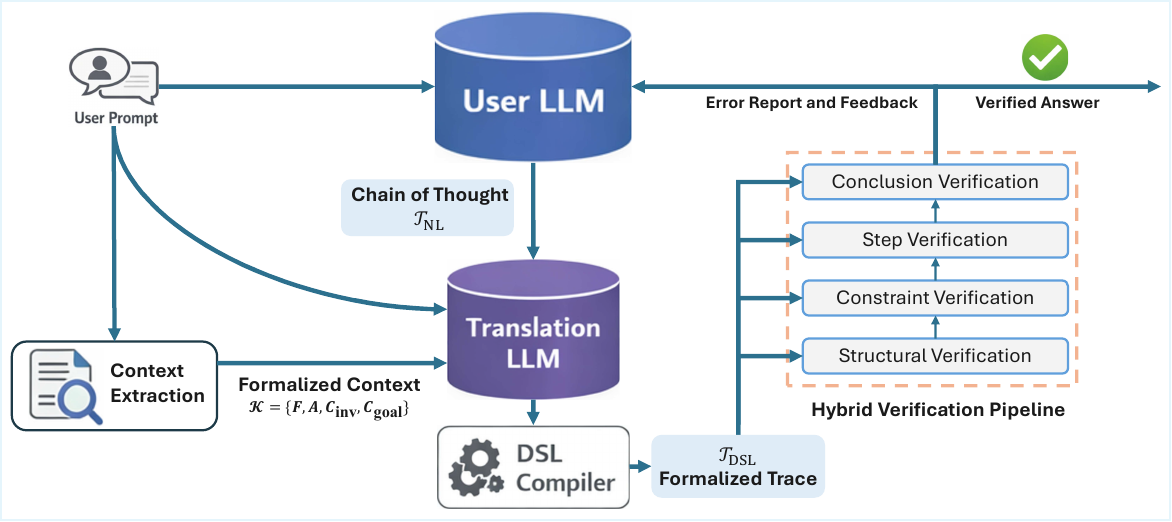}
    \caption{Illustrations of the VeryTrace Framework. The process begins with the user prompt, which is provided into the User LLM to produce a Chain of Thought, or natural language reasoning trace $\mathcal{T}_{\text{NL}}$. The Context Extraction module queries an LLM to extract formalized context $\mathcal{K}$ (initial facts $F$, assumptions $A$, invariant constraints $C_{\text{inv}}$ and goal constraints $C_{\text{goal}}$), without seeing the reasoning trace $\mathcal{T}_{\text{NL}}$. This separation prevents the model from hallucinating context to justify incorrect steps. The Chain of Thought, formalized context, alongside the user prompt are processed by a Translation LLM to produce a compilable DSL trace $\mathcal{T}_{\text{DSL}}$. This formalized trace enters the Hybrid Verification pipeline, which verifies inference structure, constraint satisfaction, computation and deduction correctness in reasoning steps, and final conclusion consistency. If the trace is invalid, the verifier generates an error report and feedback for the USER LLM to iteratively repair the reasoning trace. }
    \label{fig:verytrace}
    \vspace{-15pt}
\end{figure*}

\paragraph{A compilable DSL for reasoning traces.}
VeryTrace's Domain-Specific Language makes step dependencies explicit and maximally mechanizes quantitative content. Rather than treating ``compute $x = 2 \times 15$'' as unstructured text, the DSL represents it as \texttt{COMPUTE(x, "2 * 15", deps=[...])}, making the computational content executable and the dependency structure verifiable. For semantic inferences that resist mechanization, the DSL constrains reasoning to a small library of deduction schemas (direct entailment, modus ponens, transitivity, case analysis), ensuring even ``soft'' steps have explicit form and declared premises.

\paragraph{Hybrid verification: mechanical checks plus structured LLM audits.}
The verifier separates deterministic checks  from semantic judgments. Deterministic checks validate: (i) dependency resolution (no forward references), (ii) computational correctness (evaluate the right-hand side, verify state update), (iii) mechanizable constraint satisfaction, and (iv) conclusion consistency with declared goals. For non-mechanizable parts such as natural language constraints and underspecified commonsense, VeryTrace triggers structured LLM audits scoped to a specific step, deduction schema, and state snapshots. This division mirrors the principle: ``verify mechanically wherever possible, semantically only where necessary.''

\paragraph{Verification-driven repair.}
Step-level verification produces actionable diagnostics: not ``the answer is wrong,'' but ``Step 7 violates constraint $c$ under state $s_6$'' or ``Step 12's computation yields 42, but the state declares $x=35$.'' This enables targeted repair: an LLM can revise the problematic region rather than regenerating the entire trace. VeryTrace implements this as an iterative loop: generate CoT $\rightarrow$ formalize to DSL $\rightarrow$ verify $\rightarrow$ if invalid, return localized errors and request correction $\rightarrow$ re-verify.

\paragraph{Zero-shot transfer across domains.}
We evaluate VeryTrace on three domains chosen to stress different reasoning modes: pure symbolic manipulation (AIME 2025 competition mathematics), hybrid quantitative-semantic reasoning (LLM-BabyBench robotics planning), and relational inference over natural language (CLUTRR kinship reasoning). Across state-of-the-art open-sourced LLMs, VeryTrace improves accuracy over strong zero-shot baselines without any domain-specific training or in-context examples, demonstrating that formalized trace verification can generalize across heterogeneous reasoning types.


\paragraph{Contributions.}
In summary, this paper makes the following contributions:
\begin{enumerate}
    \item \textbf{A domain-agnostic DSL for reasoning traces.} We propose a compilable representation that makes step dependencies explicit, mechanizes quantitative content, and structures semantic inferences via universal deduction schemas.
    \item \textbf{A hybrid state-transition verifier.} We develop a step-wise verification pipeline that combines deterministic checks with structured LLM audits, propagating the reasoning state through each step.
    \item \textbf{A verification-driven repair framework.} We show how step-level error diagnostics enable targeted correction and iterative improvement of traces without training or domain-specific demonstrations.
    \item \textbf{Cross-domain evaluation.} We demonstrate consistent gains over zero-shot baselines on AIME 2025, LLM-BabyBench, and CLUTRR across multiple state-of-the-art LLMs.
\end{enumerate}

\section{Related Work}
\label{related-work}

\paragraph{Chain-of-Thought Reasoning and Verification Approaches.}
Chain-of-Thought (CoT) prompting~\cite{wei2022chain,kojima2022large} has established itself as the dominant paradigm for complex reasoning in LLMs, with extensions including Tree-of-Thoughts~\cite{yao2023tree}, Graph-of-Thoughts~\cite{besta2024graph}, and least-to-most prompting~\cite{zhou2023leasttomost}. However, these approaches fundamentally lack error detection mechanisms: a single mistake propagates uncorrected through subsequent steps. Self-consistency methods~\cite{wang2023selfconsistency} address this through majority voting over multiple samples, but require expensive generation without diagnosing \emph{why} individual traces fail. Prompt-based verification approaches attempt to mitigate this: Self-Refine~\cite{madaan2023self} and Chain-of-Verification~\cite{dhuliawala2024chain} enable iterative refinement through self-generated feedback, while Natural Program~\cite{ling2023deductive} structures reasoning as verifiable deductive steps. Recent work explores spontaneous self-correction~\cite{zhao2025boosting} and token-probability-based verification~\cite{chowdhury2025zero}. Process reward models~\cite{lightman2023lets,uesato2022solving} and step-level verifiers~\cite{li-etal-2023-making,cobbe2021training} learn to score reasoning steps but require substantial training data and remain domain-specific. VeryTrace differs fundamentally by introducing a domain-agnostic DSL that makes reasoning steps \emph{mechanically verifiable} wherever possible, providing precise error localization without requiring trained verifiers or multiple sampling.

\paragraph{Formal Methods and Solver-Integrated Reasoning.}
At the opposite end of the verification spectrum, formal proof assistants (Lean~\cite{moura2015lean}, Isabelle~\cite{paulson1994isabelle}, Coq~\cite{bertot2013coq}) offer rigorous correctness guarantees through typed calculi and tactics. Recent work has explored LLM integration with these systems: autoformalization~\cite{wu2022autoformalization} translates informal mathematics to formal statements, proof artifact co-training~\cite{han2022proof} leverages intermediate proof states, and retrieval-augmented approaches~\cite{yang2023leandojo} enable large-scale theorem proving. Draft-Sketch-Prove~\cite{jiang2023draft} and related hybrid methods~\cite{polu2020generative,trinh2024solving} use informal proofs to guide formal verification. Domain-specific systems include Safe~\cite{liu2025safe}, which translates mathematical reasoning into Lean for retrospective verification, and Typed Chain-of-Thought~\cite{perrier2025typed}, which applies type-theory principles to structure reasoning. Logic.py~\cite{kesseli2025logic} and LELMA~\cite{mensfelt2024towards} bridge natural language reasoning with constraint solvers, while SMT solvers~\cite{berman2024solving} rigorously verify constraint-heavy problems. While these approaches achieve remarkable results, they face critical limitations: (i) formalization requires expert knowledge of proof assistant syntax, (ii) the verification machinery is domain-specific, and (iii) complete formalization is often impractical for problems involving natural language constraints. VeryTrace adopts the \emph{spirit} of tactic-based proving—explicit state transitions, declared dependencies, structured inference rules—while remaining domain-agnostic through a lightweight DSL and hybrid verification strategy that mechanizes what it can and audits what it cannot.

\paragraph{Neuro-Symbolic Methods and Learned Verifiers.}
Neuro-symbolic systems~\cite{garcez2019neurosymbolic,mao2019neuro,yi2018neural} combine neural perception with symbolic reasoning, often for visual question answering, but do not address multi-step reasoning verification in language models. Work on program synthesis~\cite{chen2021evaluating,nijkamp2023codegen} and verification~\cite{solar2008program,polikarpova2016program} provides techniques for validating executable code but assumes well-defined programming language semantics. LLM-P~\cite{liu2023llmplanner} and similar planning approaches~\cite{valmeekam2022large,song2023llmplanner} integrate LLMs with classical planners requiring domain-specific planning languages (PDDL). Learned verification approaches train specialized models to assess reasoning quality: Math-Shepherd~\cite{wang2024math} and ThinkPRM~\cite{khalifa2025process} advance Process Reward Models (PRMs) where trained verifiers score individual steps, DiVeRSe~\cite{li-etal-2023-making} employs step-aware verifiers with voting mechanisms, and S$^2$R~\cite{ma-etal-2025-s2r} uses reinforcement learning to train critics for self-verification. Recent work~\cite{zhao2025verifying} proposes verification via computational graph analysis. Code-based verification methods~\cite{zhou2023solving} demonstrate the value of executable representations for mathematical reasoning but remain specialized to arithmetic domains. In contrast to these methods, which typically require extensive training data, domain-specific solvers, or full formalization, VeryTrace generalizes by treating reasoning traces as programs expressed in a universal DSL, enabling mechanized verification of quantitative content while using structured LLM audits for semantic content that resists formalization. This hybrid approach allows VeryTrace to transfer zero-shot across heterogeneous domains—from pure mathematics to robotics planning to relational reasoning—without domain-specific compilation, solvers, or training data, filling a critical gap between brittle end-to-end verification and impractical full formalization.

\section{DSL for Reasoning Trace Formalization}
\label{sec:dsl}

VeryTrace formalizes free-form reasoning as a sequence of state transitions governed by a \emph{compilable} DSL. Such a formalization (a) makes step dependencies explicit, (b) mechanizes as much content as possible into executable expressions, and (c) structures the remaining semantic inferences via a small library of universal deduction schemas.

\subsection{Reasoning State}
\label{sec:reasoning-state-def}
We formalize a reasoning trace as state transitions over an explicit \emph{reasoning state space}. 

Let $\mathcal{S}$ denote a reasoning state space. A reasoning state $s \in \mathcal{S}$ is a partial binding
\[
s:\;\mathcal{V} \to \mathcal{D},
\]
mapping variable names $\mathcal{V}$ (e.g., \texttt{money}, \texttt{position}) to typed values in a domain $\mathcal{D}$ (e.g., numbers, strings, tuples, lists). The DSL state is designed to be domain-agnostic.

\subsection{Formalizing Problem Context}
\label{sec:context-def}
We explicitly decouple the problem context from the reasoning trajectory. We formalize the problem context as $\mathcal{K} = (F, A, C_{\text{inv}}, C_{\text{goal}})$, consisting of initial facts, assumptions, and constraints.

\textbf{Initial Facts ($F$).} We define $F = \{f^1, f^2, \dots, f^M\}$ as a set of $M$ indisputable premises derived explicitly from the problem statement. Each initial fact $f^i$ is a proposition that seeds the first reasoning state $s_1$ (e.g., \textit{``Betty initially has \$50''}, \textit{``Robot starts at $(3,4)$ facing East''}). These serve as the roots of the dependency graph, providing  ground truth for anchoring subsequent inferences.

\textbf{Assumptions ($A$).} We define $A = \{a^1, a^2, \dots, a^K\}$ as a set of $K$ domain knowledge from the context for all reasoning steps. Some assumptions may be explicitly stated (e.g., \textit{"Assume all divisions in this context to be integer divisions"}), while others are implied (e.g., \textit{"moving North increases the y-coordinate in standard coordinate systems"}). By formalizing these as explicit DSL assumptions, we reduce the LLM's tendency to hallucinate convenient facts when justifying incorrect steps.

\textbf{Constraints ($C$).} To enforce safety and reliability, we define constraints $C=\{c_1,\ldots,c_N\}$ as  conditions that must hold for valid reasoning. Each constraint $c_i \in C$ is modeled as a Boolean predicate over reasoning states:
\[
c_i:\mathcal{S}\to\{\text{true},\text{false}\}.
\]
where $c_i(s_t)$ evaluates to true if and only if the reasoning state $s_t$ satisfies the constraint.

VeryTrace distinguishes between two constraint scopes:
\begin{itemize}
    \item \textbf{Invariant constraints} $C_{\mathrm{inv}}$ must hold at \emph{every} state along the trace (e.g., ``the agent never visits a forbidden cell'').
    \item \textbf{Goal constraints} $C_{\mathrm{goal}}$ must hold only at the \emph{final} state (e.g., ``the agent reaches the target'', ``the final numerical answer matches the required format'').
\end{itemize}
This separation enables more precise definitions of reasoning constraints.

\subsection{Typed Reasoning Step}
\label{sec:reasoning-step-def}
We formalize a natural language reasoning trace $\mathcal{T}_{\text{NL}}$ as a DSL trace $\mathcal{T}_{\text{DSL}}$ in the form of a sequence of state transitions through typed reasoning steps. Each reasoning step consumes a set of premises, advances the reasoning state and produces an intermediate claim. 

Let $\sigma_t$ be a reasoning step that advances a reasoning state $s_t$ to $s_{t+1}$ and produces an intermediate claim $\varphi_{t+1}$. A \emph{claim} $\varphi_{t+1}$ is an atomic assertion or proposition derived from the step $\sigma_t$. We further denote the set of premises referenced by $\sigma_t$ as $P_t$. Each \emph{premise} $p \in P_t$ is either (i) an initial fact $f \in F$, (ii) an assumption $a \in A$ from the context, or (iii) a prior claim $\varphi_k,$ where $k < t$.

As such, a valid reasoning state transition is defined as:
\begin{equation}
    \sigma_t(\texttt{type}_t, s_t, P_t) = (s_{t+1}, \varphi_{t+1})
\end{equation}
where $\texttt{type}_t$ indicates the step's inference type, determining its verification method. VeryTrace defines three inference types for reasoning steps: COMPUTE, ASSUME, DEDUCE, and CONCLUDE. 

\textbf{COMPUTE step.} The step must provide one or more executable assignment expressions to some state variables (e.g., $x = \sin{\theta} + k$)  alongside an \texttt{update} field defining the state change. This enables deterministic verification: the verifier checks whether $s_{t+1}$ matches evaluating  $\sigma_t$'s expression under $s_t$,  eliminating arithmetic hallucinations.

\textbf{ASSUME step.} The step declares a premise from the context $\mathcal{K}$. It links the trace explicitly to the static context.

\textbf{DEDUCE step.} The step performs  logical inference. To structure semantic reasoning, we require \textsc{Deduce} steps to instantiate \textit{Deduction Schemas} from a fixed library. This enforces a structured representation of natural language deductions, enabling rigorous audits.


The VeryTrace deduction library includes:

\begin{itemize}
    \item  \textit{Modus Ponens}: Given $A \to B$ and $A = \text{true}$, infer $B$.
    \item  \textit{Conjunction}: Given $A = \text{true}$ and $B = \text{true}$, then $A \land B = \text{true}$.
    \item \textit{Transitivity}: Given $x = y$ and $y = z$, infer $x = z$.
    \item \textit{Direct Deduction}: To ensure flexibility, this schema is an unstructured fallback when other structured schemas are too restrictive. 
\end{itemize}

\textbf{CONCLUDE step.} This step produces the final reasoning state and declares the final conclusion, denoted as $\omega$.

Finally, we formally define a DSL trace $\mathcal{T}_{DSL}$ as a trajectory of reasoning state transitions with a sequence of intermediate claims, seeded with the problem context $\mathcal{K} = (F, A, C_{\text{inv}}, C_{\text{goal}})$, arriving at a final conclusion $\omega$:

\begin{equation}
    \mathcal{T}_{\text{DSL}} := \Big(\mathcal{K},\; \{s_t\}_{t=1}^{T},\; \{\sigma_t\}_{t=1}^{T-1},\; \{\varphi_t\}_{t=2}^{T}, \omega\Big)
    \label{eq:dsl-def}
\end{equation}

$\mathcal{T}_{\text{DSL}}$ is \emph{compilable} because (i) computational expressions and executable constraints can be parsed and evaluated against the evolving state, and (ii) semantic components have a standardized structure (premises, claims, schema arguments) enabling structured audits. In the following section, we present the verification pipeline, which determines whether a given trace $\mathcal{T}_{\text{DSL}}$ is valid.

\section{Structured Verification}
\label{sec:verifier}

\begin{figure*}[tb]
    \centering
    \includegraphics[width=\linewidth]{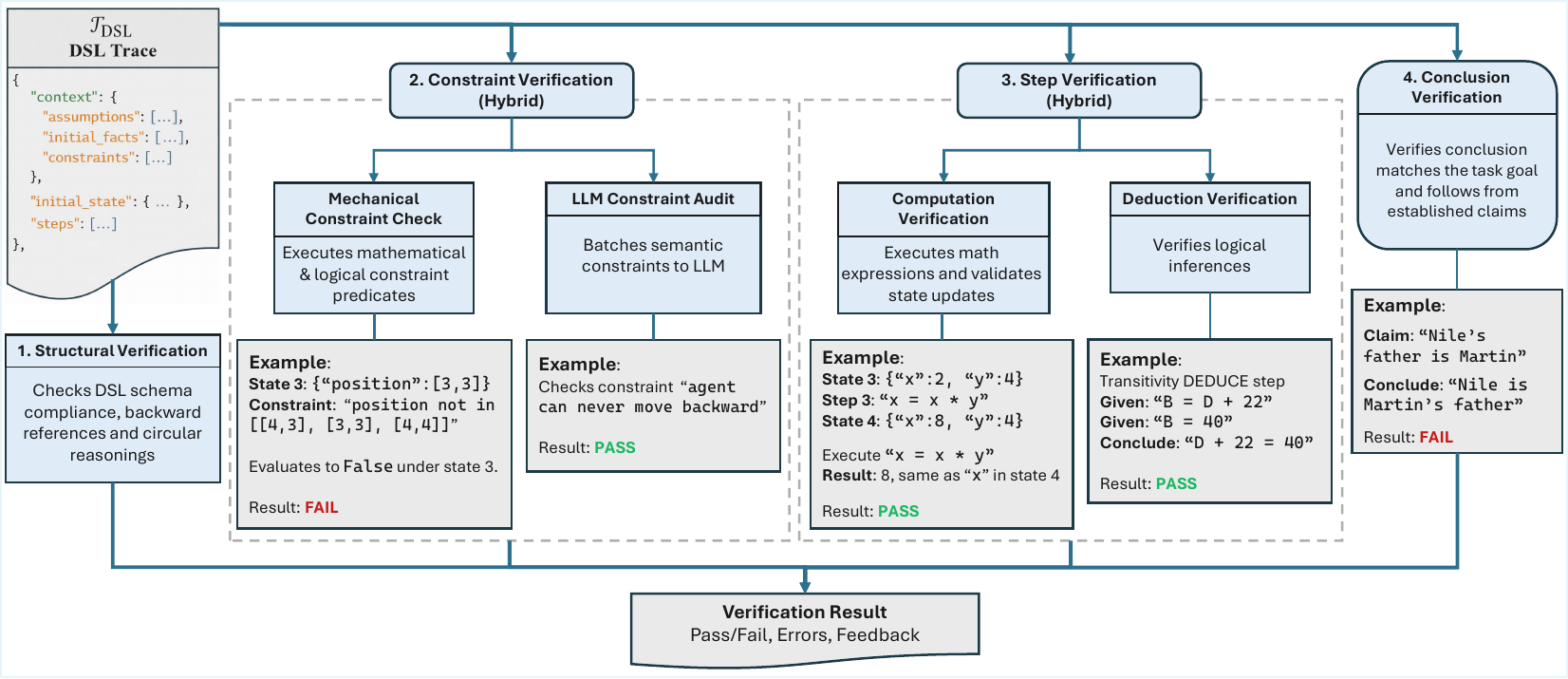}
    \caption{Illustrations of Structured Verification Pipeline. Given a compiled DSL trace $\mathcal{T}_{\text{DSL}}$, the verification pipeline applies four stages of logical verifications. \textbf{(1) Structural Verification} checks for backward references and circular dependencies. \textbf{(2) Constraint Verification} verifies each reasoning state on constraint satisfactions. \textbf{(3) Step Verification } verifies the computation in COMPUTE step, and logical deduction validity from DEDUCE and ASSUME steps. Finally, \textbf{(4) Conclusion Verification} verifies whether the final conclusion follows logically from the established claims. }
    \label{fig:verification-pipeline}
    \vspace{-15pt}
\end{figure*}

We now present a verification pipeline operating on the formalism defined in \secref{sec:dsl}. As illustrated in \figref{fig:verytrace}, the pipeline implements a \textit{Generate-Verify-Repair} loop: the user LLM generates a reasoning trace, which is compiled into the VeryTrace DSL, verified for correctness, and if invalid, returned to the model with precise, step-level diagnostics for repair. This section details two critical components: bias-mitigating DSL conversion and hybrid verification logic.

\subsection{Two-stage DSL conversion}
\label{sec:conversion}
The conversion from the prompt $Q$ and the natural language reasoning trace $\mathcal{T}_{\text{NL}}$ to a DSL trace $\mathcal{T}_{\text{DSL}}$ is itself  critical: verification quality depends on correctly extracting problem context and reasoning step structure. If the translation model conditions on the reasoning trace, it may overlook errors to maintain consistency with the provided reasoning \cite{Holtzman2020The}. For example, if a trace violates a constraint from the prompt, a naive translator might omit that constraint from the DSL to make the trace appear valid. Therefore, we use a 2-stage conversion:

\paragraph{Stage 1: context extraction.}
A \emph{context-extraction LLM} reads only the user prompt $Q$ and produces a formalized problem context $\mathcal{K}$. Crucially, it does \emph{not} see the trace $\mathcal{T}_{\text{NL}}$, preventing missing or hallucinated context from propagating to the DSL representation.

\paragraph{Stage 2: trace translation.}
A \emph{translation LLM} then receives $(Q,\mathcal{K}, \mathcal{T}_{\text{NL}})$ and produces the DSL trace $\mathcal{T}_{\text{DSL}}$. The translation prompt encourages maximal mechanization: constraints and computations should use executable expressions (e.g., arithmetic, equality/inequality, membership tests) whenever possible, enabling deterministic verification over LLM audits.

Our Ablation study (\secref{sec:eval-ablation}) suggests that this separation benefits tasks involving longer reasoning chains and mathematical reasoning where explicitly stated context is needed. We further evaluate the quality of the translation and its impact to the downstream performance in Appendix~\ref{app:translation-ablation}.

\subsection{Hybrid verification pipeline}
Given a compiled $\mathcal{T}_{\text{DSL}}$ from Eq.~\ref{eq:dsl-def}, the verifier performs a sequence of verifications that combines mechanical validations with structured LLM audits only where needed. Mechanical validations provide guarantees and reliability, while  LLM-audits enable greater flexibility. The complementary nature of mechanical and LLM-based verification are further analyzed and highlighted in Appendix~\ref{app:mechanical-v-llm-ablation} and~\ref{app:split-mechanical-llm}.

\subsubsection{structural Verification ($V_{\mathrm{str}}$)}
Before evaluating content, we verify the structural integrity of the DSL trace. This verification first checks that each step contains the required fields for its inference type (e.g., COMPUTE steps provide executable expressions; DEDUCE steps instantiate deduction schemas). Further, we verify that every premise $p \in p_t$ referenced in step $\sigma_t$ refers to a claim established at a prior step $t$. This mechanically detects forward \& backward references and circular dependencies. We denote this verification result as $V_{pre}(\mathcal{T}_{\text{DSL}}) \in \{\text{true, false}\}$.

\subsubsection{constraint Verification ($V_{\mathrm{cst}}$)}
\label{sec:constraint-verification}
We verify that the reasoning trajectory respects the problem constraints defined in the context. Given the reasoning trace $\mathcal{T}_{\text{DSL}} := (\mathcal{K},\; \{s_t\}_{t=1}^{T},\; \{\sigma_t\}_{t=1}^{T-1},\; \{\varphi_t\}_{t=2}^{T}, \omega)$, and let $C_{\mathrm{inv}}$ and
$C_{\mathrm{goal}}$ be invariant and goal constraints from context $\mathcal{K}$. We strictly enforce that all invariant constraints hold at every state, and that goal constraints hold at the final state. Formally, we define the constraint verification result $V_{cst}(\mathcal{T}_{\text{DSL}}) \in \{\text{true}, \text{false}\}$ as:
\begin{equation}
    V_{cst}(\mathcal{T}_{\text{DSL}}) = \left(\bigwedge_{t=1}^{T} \bigwedge_{c \in C_{inv}} c(s_t) \right) \land \left( \bigwedge_{c \in C_{\text{goal}}} c(s_{T}) \right)
    \label{eq:constraint-check}
\end{equation}

Not all constraints admit an executable predicate form (e.g., natural-language constraints like
\textit{``in the game, rule5 takes higher precedence over rule1''}). Hence, to evaluate Eq.~\ref{eq:constraint-check}, we employ a hybrid strategy: If a constraint $c$ is expressed as a boolean predicate (e.g., $position \neq [2,3]$), it is evaluated deterministically against the state $s_t$. For all remaining constraints $c$ that represent semantic constraints, we batch these constraints into a prompt and query an Audit LLM. This audit LLM verifies whether the state $s_t$ satisfies the natural language specification of $c$.

\subsubsection{computation Verification}
\label{sec:compute-check}

For COMPUTE steps, we perform deterministic verification. The verifier extracts the executable math expressions (possibly multiple), evaluates them using the values from the prior state $s_t$, and compares the results to the claimed update in $s_{t+1}$.  Discrepancies trigger verification failures. This verification detects arithmetic hallucinations, ensuring mathematically sound quantitative updates.

\subsubsection{assumption and deduction Verification}
\label{sec:deduce-check}
For \textsc{Assume} and \textsc{Deduce} steps, mechanical verification is insufficient as they often involve natural language entailment. Instead, we use \textbf{Structured LLM Audits}. For each deduction step, we construct a verification prompt containing  explicit premises $p_t$, the declared deduction schema, and the derived claim $\phi_{t+1}$. The Audit LLM is tasked with a single, highly focused judgment: \textit{Does the claim logically follow from the premises using the specified rule?}

For ASSUME steps, the verifier audits whether the claimed assumption is consistent with problem context K.

\paragraph{Step-wise validity.}
Let $V_{\mathrm{step}}(\mathcal{T}_{\text{DSL}}) \in \{\text{true}, \text{false}\}$ denote the step-wise validity verification result of the trace $\mathcal{T}_{\text{DSL}}$, defined as:
\begin{equation}
\label{eq:step}
V_{\mathrm{step}}(\mathcal{T}_{\text{DSL}}) \;=\; \bigwedge_{t=1}^{T-1} V_t,
\end{equation}
where $V_t$ is computed by (\secref{sec:compute-check}) deterministic execution and verification for COMPUTE steps,  and (\secref{sec:deduce-check}) structured audits for DEDUCE and ASSUME steps.

\subsubsection{conclusion verification ($V_{\mathrm{conc}}$)}
Local step checks do not guarantee that the final conclusion $\omega$ follows from established claims. We therefore employ an Audit LLM to verify that the final conclusion $\omega$ follows from established reasoning. The audit prompt includes the problem context $\mathcal{K} = (F, A, C_{\text{inv}}, C_{\text{goal}})$, and the sequence of established claims $\{\varphi_t\}$. The LLM  assumes all prior claims are valid and all constraints  are satisfied, then verifies whether the conclusion $\omega$ follows. This detects cases where intermediate reasoning is sound but the final answer is inconsistent.

Finally, the validity of a given DSL trace $\mathcal{T}_{\text{DSL}}$ is the conjunction of all verification components:

\begin{equation}
    \begin{aligned}
V(\mathcal{T}_{\text{DSL}}) \;=\;&
V_{\mathrm{str}}(\mathcal{T}_{\text{DSL}})\wedge V_{\mathrm{cst}}(\mathcal{T}_{\text{DSL}}) \\
&\wedge V_{\mathrm{step}}(\mathcal{T}_{\text{DSL}})\wedge V_{\mathrm{conc}}(\mathcal{T}_{\text{DSL}}).
\end{aligned}
\end{equation}

\subsection{Verification-driven repair}
When $V(\mathcal{T}_{\text{DSL}})=\text{false}$, the verifier returns structured diagnostics containing: step index, error type or violated constraint, and explanation. Because errors are localized (e.g., \textit{``State 7 violates invariant constraint $c$''} or \textit{``Step 12's computation disagrees with its claimed values''}), the user LLM can revise the problematic region instead of regenerating the full trace. The verification feedback loop terminates when a valid trace is produced or a maximum iteration count $R_{\text{max}}$ is reached. The overall algorithm is presented in Alg.~\ref{alg:feedback-loop}.

\begin{algorithm}
    \small
    \caption{VeryTrace Verification Feedback Loop} 
    \label{alg:feedback-loop}
    \begin{algorithmic}[1]
    \STATE {\bfseries Input:} user prompt $Q$, max iteration $R_{\text{max}}$
    \STATE {\bfseries Output:} Verified answer $\omega$
    \STATE $\omega, \mathcal{T}_{\text{NL}} \leftarrow $ LLM$(Q)$ 

    \FOR{$i=1$ {\bfseries to} $R_{\text{max}}$}
    \STATE $\mathcal{K} \leftarrow $ ExtractContext$(Q)$
    
    \STATE $\mathcal{T}_{\text{DSL}} \leftarrow $ TranslateDSL$(Q, \mathcal{K}, \mathcal{T}_{\text{NL}})$

    \STATE $V_{\mathrm{pre}}(\mathcal{T}_{\text{DSL}}) \leftarrow$ Structural Verification

    \STATE $V_{\mathrm{cst}}(\mathcal{T}_{\text{DSL}}) \leftarrow$ Constraint Verification

    \STATE $V_{\mathrm{step}}(\mathcal{T}_{\text{DSL}}) \leftarrow$ Step-wise Validity Verification

    \STATE $V_{\mathrm{conc}}(\mathcal{T}_{\text{DSL}}) \leftarrow$ Conclusion Verification

    \STATE $\begin{aligned}
V(\mathcal{T}_{\text{DSL}}) \;=\;&
V_{\mathrm{str}}(\mathcal{T}_{\text{DSL}})\wedge V_{\mathrm{cst}}(\mathcal{T}_{\text{DSL}}) \\
&\wedge V_{\mathrm{step}}(\mathcal{T}_{\text{DSL}})\wedge V_{\mathrm{conc}}(\mathcal{T}_{\text{DSL}}).
\end{aligned}$

    \IF{$V(\mathcal{T}_{\text{DSL}}) == \text{true}$}
    \STATE \textbf{return} $\omega$
    \ELSE 
    \STATE $\epsilon \leftarrow$ generate logical error feedback
    \STATE $\omega, \mathcal{T}_{\text{NL}} \leftarrow $ LLM$(Q, \epsilon)$
    \ENDIF
    \ENDFOR

    \STATE \textbf{return} $\omega$
    \end{algorithmic}
\end{algorithm}
\vspace{-3mm}
\newcommand{\llama}{Llama-3.3-70B-Instruct}
\newcommand{\qwentwo}{Qwen3-Next-80B-A3B-Instruct}
\newcommand{\qwenthink}{Qwen3-Next-80B-A3B-Thinking}

\section{Evaluation}
\label{sec:eval}

We evaluate VeryTrace to assess whether formalizing natural language reasoning and applying hybrid verification improve the reliability of LLM reasoning. We aim to answer three key questions: (1) Does VeryTrace improve performance across diverse reasoning domains compared to strong zero-shot baselines? (2) Does the structured verification framework prevent error propagation in long-horizon reasoning tasks? (3) Are the deterministic mechanical checks and the two-stage DSL conversion process necessary for accurate verification?

\subsection{Experimental Setup}
\label{sec:eval-setup}

\textbf{Models.} We employ state-of-the-art open-weights models for both the reasoning agents and the underlying components of the VeryTrace pipeline. For the user LLM, we test \llama{} (L3.3-70b-I), \qwentwo{} (Q3-80b-I), and the reasoning-specialized \qwenthink{} (Q3-80b-T). For the VeryTrace pipeline (Context Extraction, Translation, and Audit), we uniformly utilize Qwen3-Next-80B-Thinking to ensure high-quality formalization and semantic auditing. For all experiments, we set the maximum repair budget $R_{max}=5$. All methods are evaluated in a zero-shot setting without task-specific fine-tuning.

\paragraph{Baselines.}
We compare against two established prompt-based verification baselines: (1) \textit{Chain-of-Verification (CoVe)}~\cite{dhuliawala2024chain}, a prompt-based method where the model drafts verification questions to self-correct hallucinations; and (2) \textit{Natural Program (NP)}~\cite{ling2023deductive}, which encourages the generation of deductive steps verified against problem premises. We also include the  base model (Vanilla) as a baseline, which generates answers via standard Chain-of-Thought prompting without verification.

\textbf{Domains and Metrics.} We select three datasets to stress-test different reasoning modalities. First, for symbolic and quantitative reasoning, we use 30 problems from the \textbf{AIME 2025} competition. This domain  tests arithmetic precision and symbolic manipulation in  reasoning traces. Second, for the mixed quantitative-semantic reasoning domain, we utilize  planning tasks from \textbf{LLM-BabyBench}~\cite{choukrani2025llm}. Models must plan action sequences that navigate a robot through a gridworld from an initial position to a goal while avoiding obstacles. This task tests the pipeline's ability to maintain state consistency (position, direction) and respect constraints over long horizons.
We run LLM-BabyBench under three difficulty levels (\textit{small/medium/large}) that vary world size and thus plan horizon. We evaluate 100 trials per difficulty (300 total). Finally, for  semantic reasoning, we use the test split (1048 queries) of the \textbf{CLUTRR} benchmark~\cite{sinha2019clutrr} to test the semantic inference of kinship relations from given stories.

For AIME 2025 and CLUTRR, we report \textbf{accuracy} via exact match against ground truth answers. For LLM-BabyBench Planning, where multiple feasible paths  exist, we report the \textbf{success rate} by executing the predicted action sequence in a simulated grid world using Gymnasium~\cite{towers2024gymnasium} and checking whether it reaches a terminal state that satisfies the task completion condition (agent ends adjacent to the target cell while facing it).

\subsection{Main Results and Discussion}
\label{sec:eval-main}

\begin{table*}[t]
\centering
\caption{Performance comparison of our method against baselines across three domains and three models.}
\label{tab:evaluation_results}
\begin{tabular}{l ccc ccc ccc}
\toprule
& \multicolumn{3}{c}{\textbf{AIME 2025}} 
& \multicolumn{3}{c}{\textbf{LLM-BabyBench Planning}} 
& \multicolumn{3}{c}{\textbf{CLUTRR}} \\

\cmidrule(lr){2-4} \cmidrule(lr){5-7} \cmidrule(lr){8-10}

\textbf{Method} & \mhead{L3.3-70b-I} & \mhead{Q3-80b-I} & \mhead{Q3-80b-T} & \mhead{L3.3-70b-I} & \mhead{Q3-80b-I} & \mhead{Q3-80b-T} & \mhead{L3.3-70b-I} & \mhead{Q3-80b-I} & \mhead{Q3-80b-T} \\
\midrule

Vanilla & 3.33 & 53.33 & 83.33 & 8.67 & 40.00 & 60.33 & 36.67 & 27.00 & 36.00 \\
CoVe & 3.33 & 73.33 & 86.67 & 13.67 & 50.00 & 70.00 & 44.00 & 32.67 & \textbf{70.33} \\
NP & 6.67 & 73.33 & 83.33 & 11.00 & 45.00 & 68.33 & 40.67 & 41.00 & 58.33 \\
VeryTrace & \textbf{26.67} & \textbf{80.00} & \textbf{90.00} & \textbf{36.33} & \textbf{78.00} & \textbf{89.33} & \textbf{61.67} & \textbf{48.67} & 70.00 \\

\bottomrule
\end{tabular}
\end{table*}

Table~\ref{tab:evaluation_results} reports performance across all domains and models. The per-difficulty breakdown for LLM-BabyBench Planning is presented in Fig.~\ref{fig:babybench-eval}.

VeryTrace generally achieves the highest accuracy, demonstrating that formalizing the reasoning process yields gains regardless of the underlying model architecture.

A key finding is the performance impact on reasoning-specialized models. VeryTrace significantly improves the performance of \qwenthink{} (Q3-80b-T). While Q3-80b-T achieves a strong baseline of 83.33\% on AIME 2025, applying VeryTrace boosts this to 90.00\%. This result suggests that even specialized models for complex reasoning tasks benefit from the externalized, rigid reasoning state tracking provided by our DSL. The deterministic checks in VeryTrace ($V_{cst}$ and mechanical \texttt{COMPUTE} verification) potentially act as a hard runtime monitor that purely neural self-correction cannot replicate.

Furthermore, these gains largely generalize across domains, but the margin over CoVe shrinks on CLUTRR with Q3-80B-T. We hypothesize that given the CLUTRR benchmark is purely semantic reasoning, VeryTrace has fewer opportunities to exploit mechanical verification. This suggests that VeryTrace overall generalizes across domains, with the strongest benefits likely in math and symbolic reasoning.

\subsection{Reasoning in Varying Planning Horizons}
\label{subsec:robustness}

\begin{figure}
    \centering
    \includegraphics[width=\linewidth]{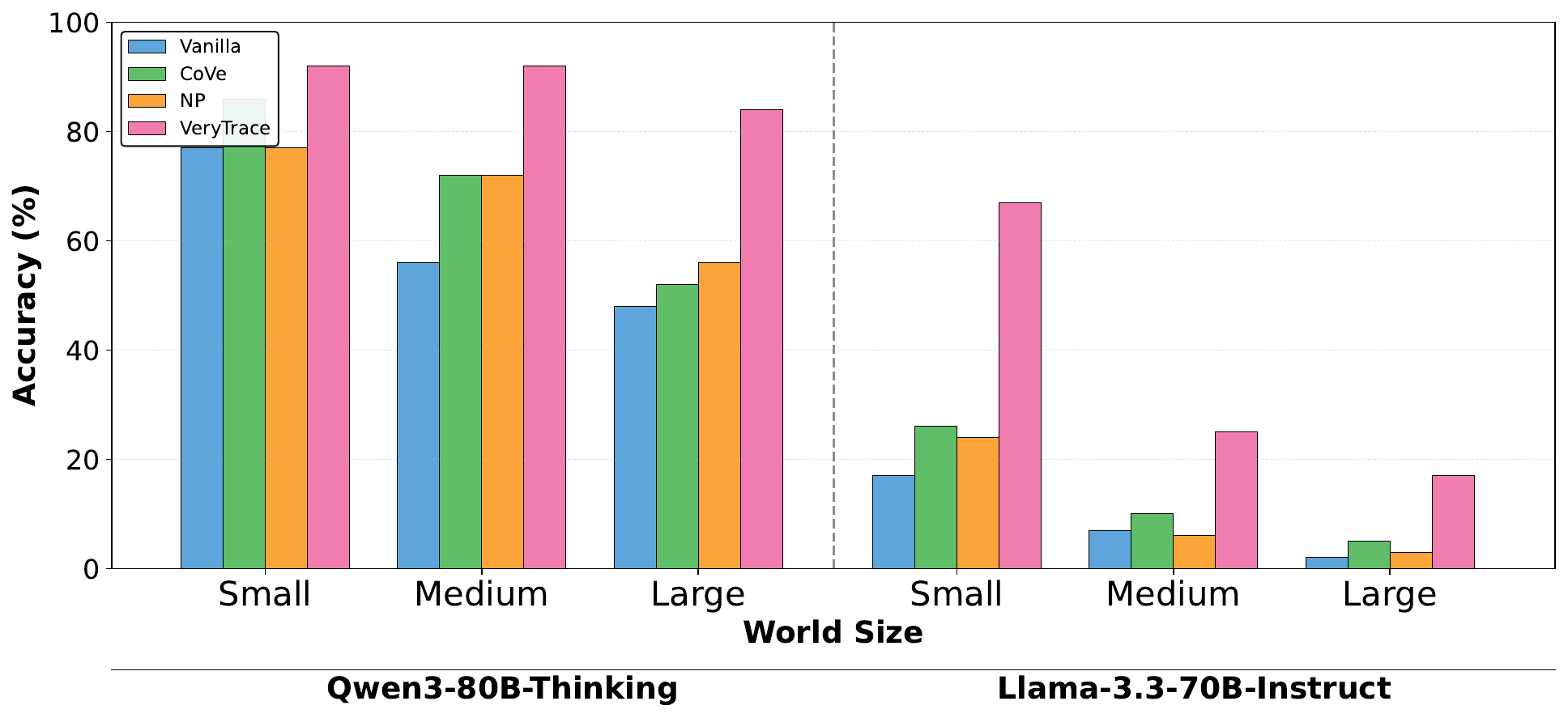}
    \caption{LLM-BabyBench Planning Results Across World Sizes}
    \label{fig:babybench-eval}
    \vspace{-10pt}
\end{figure}

We further analyze VeryTrace under increasing reasoning horizons by evaluating LLM-BabyBench as the world size (and hence planning horizon) scales from \textit{Small} to \textit{Large}. Figure~\ref{fig:babybench-eval} reports the corresponding success rates. We present results for two models, \qwenthink{} and \llama{}; results for \qwentwo{} are deferred to Appendix~\ref{app:babybench-eval}.

As expected, performance for all methods degrades as difficulty increases. VeryTrace outperforms all other methods in all sizes, indicating its ability for long-horizon reasoning.

For experiments using \qwenthink{} (Figure~\ref{fig:babybench-eval} left), we implement Vanilla, CoVe, and Natural Program using the same underlying model as VeryTrace, since VeryTrace also uses \qwenthink{} for all components. The resulting gains therefore reflect the DSL formalization and structured verification, rather than benefits from selecting different audit LLMs.

\subsection{Ablation Study: Two-stage Translation}
\label{sec:eval-ablation}

VeryTrace uses a two-stage DSL translation to reduce translator bias (\secref{sec:conversion}). To test whether this separation matters, we compare against a variant with \textit{direct} translation (Direct), which collapses the two steps into a single call that directly maps $(Q,T_{\mathrm{NL}})\rightarrow T_{\mathrm{DSL}}$, while keeping the rest of the VeryTrace verification pipeline unchanged. We evaluate \textit{Two-Stage} vs.\ \textit{Direct} on the AIME 2025 dataset, as well as two 150-instance subsets: LLM-BabyBench Planning (Small) and CLUTRR. The results are shown in Table~\ref{table:ablation-study}.

\begin{table}
\centering
\begin{tabular}{lccc}
    \toprule
     & AIME 2025 & Planning & CLUTRR \\
     \midrule
    \textit{Two-Stage} & \textbf{90.00} & \textbf{94.67} & 69.33 \\
    \textit{Direct} & 86.67 & 88.44 & \textbf{70.67}  \\
\bottomrule
\end{tabular}
\caption{\textit{Direct} translation vs. \textit{Two-Stage} translation}
\label{table:ablation-study}
\vspace{-20pt}
\end{table}



Table~\ref{table:ablation-study} suggests that two-stage translation is beneficial in AIME 2025 and Planning, but underperforms \textit{Direct} on CLUTRR. We hypothesize this difference reflects the fact that AIME and Planning are constraint and computation-heavy, where isolating $\mathcal{K}$ before seeing $T_{\mathrm{NL}}$ can better preserve a strict problem specification for verification. In contrast, CLUTRR relies more on purely semantic inference with fewer explicit constraints to extract; in this setting, a separate context pass may instead introduce noise. We leave a broader ablation across additional semantic datasets and larger CLUTRR slices to future work.

\section{Conclusion}
\label{conclusion}

We presented \textsc{VeryTrace}, a reasoning formalization and structured verification framework that compiles natural-language reasoning traces into explicit state transitions and verifiable claims. VeryTrace combines deterministic checks with LLM audits, producing localized error reports for iterative repair. Across AIME 2025, LLM-BabyBench planning, and CLUTRR benchmarks, VeryTrace improves performance over strong prompt-based baselines and yields gains for reasoning-specialized models, suggesting that formalizing reasoning traces and structured verifications provide a robust framework for validating logical reasoning.


\section*{Impact Statement}

This work aims to improve the reliability and verifiability of Large Language Model reasoning, reducing hallucinations and error propagation in multi-step reasoning tasks. By enabling more transparent and auditable reasoning traces, VeryTrace aims to enhance trust and safety in LLM deployment for potentially high-stakes applications.

The computational overhead of our verification pipeline may limit accessibility, potentially concentrating benefits among well-resourced actors. Future work should address efficiency to democratize access to verified reasoning.


\bibliography{example_paper,verytrace_related_work}
\bibliographystyle{icml2026}

\newpage
\appendix
\onecolumn

\lstdefinestyle{promptstyle}{
  basicstyle=\ttfamily\scriptsize,
  breaklines=true,
  breakatwhitespace=false,
  columns=fullflexible,
  frame=single,
  framerule=0.4pt,
  xleftmargin=0.6em,
  xrightmargin=0.2em,
  aboveskip=0.6em,
  belowskip=0.6em,
  showstringspaces=false
}

\section{Reasoning in Varying Planning Horizons with \qwentwo{}}
\label{app:babybench-eval}
\begin{figure}
    \centering
    \includegraphics[width=\linewidth]{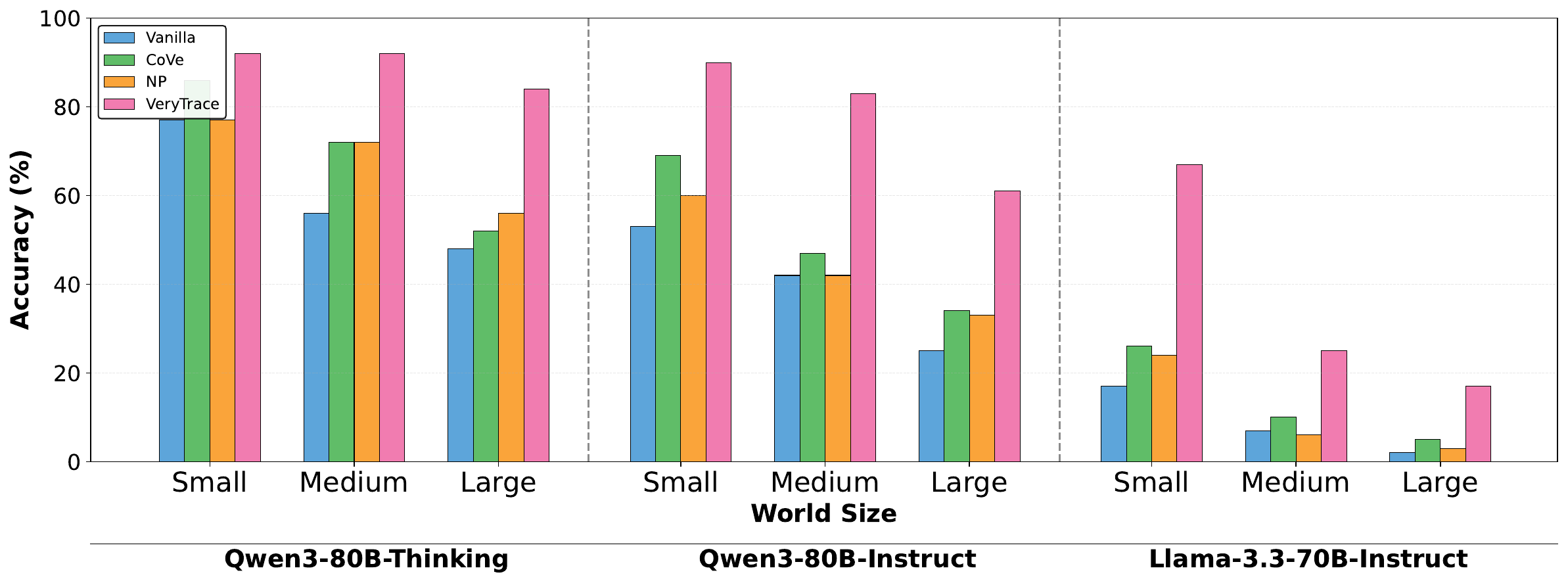}
    \caption{LLM-BabyBench Planning Results Across World Sizes}
    \label{fig:babybench-eval-all}
    \vspace{-10pt}
\end{figure}

Here we present the evaluation results under LLM-BabyBench Planning in varying planning horizons with all three tested LLM models: \qwenthink{}, \qwentwo{}, and \llama{}. The results are presented in Fig.~\ref{fig:babybench-eval-all}, which is consistent with the analysis from Sec.~\ref{subsec:robustness}.

In particular, Fig.~\ref{fig:babybench-eval-all} shows that while all methods degrade as the planning horizon grows, VeryTrace remains the strongest method across all world sizes. The advantage is especially clear for weaker planning backbones (\qwentwo{} and \llama{}), where VeryTrace retains substantially higher success rates under Medium and Large worlds.

\section{Ablation Study on Mechanical Verifications}
\label{app:mechanical-v-llm-ablation}
To further analyze the effectiveness of the hybrid verification pipeline, namely the inclusion of mechanical verifications in comparison with purely an LLM-as-a-judge pipeline, we developed a variant of VeryTrace, termed VeryTrace-LLM, where all the mechanical verifications are replaced by LLM-audits. Specifically, this involves replacing constraint verification (Sec.~\ref{sec:constraint-verification}) and COMPUTE step verification (Sec.~\ref{sec:compute-check}) with LLM audits. We ran a ablation study between VeryTrace and VeryTrace-LLM using the \qwenthink{} model. We ran the ablation study on AIME 2025, BabyBench Planning (100 trials each for small, medium, and large), and CLUTRR. The results are presented in Table~\ref{table:ablation-study-llm}.

\begin{table}
\centering
\begin{tabular}{lccccc}
    \toprule
     & AIME 2025 & Planning-S & Planning-M & Planning-L & CLUTRR \\
     \midrule
    VeryTrace & \textbf{90.00} & 98.00 & \textbf{91.00} & \textbf{82.00} & \textbf{69.4}\\
    VeryTrace-LLM & 83.3 & \textbf{99.00} & 84.00  & 77.00 & 67.9\\
\bottomrule
\end{tabular}
\caption{Ablation study between VeryTrace and purely LLM-audit variant VeryTrace-LLM}
\label{table:ablation-study-llm}
\end{table}

As shown, VeryTrace generally outperforms the VeryTrace-LLM variant, especially under the AIME and BabyBench Planning benchmarks and with increased reasoning horizons. This suggests that the mechanical verifications provide reliable audits overall, and particularly for planning and math related tasks.

\section{Assessment on Translation Quality Impact}
\label{app:translation-ablation}
To directly probe translation quality and its impact on downstream verification performance, we added a faithfulness analysis. Given a natural-language CoT and its translated DSL trace, we ask an audit LLM to rate translation faithfulness on a 1–10 scale using an explicit rubric (see App.~\ref{app:prompt-context-extraction}). We then compare faithfulness for cases where verification is correct vs. incorrect. The correctness of verification is obtained by comparing the verification output against the ground truth correctness of the reasoning. For this analysis, we ran VeryTrace on 400 trials from BabyBench Planning (Large). The results are in Table~\ref{table:ablation-translation-faithfulness}.

\begin{table}
\centering
\begin{tabular}{lccc}
    \toprule
    Verification Outcome & Mean Faithfulness & STD Faithfulness \\
     \midrule
    Correct Verification & 7.72 & 0.5712 \\
    Incorrect verification & 7.55 & 0.809 \\
\bottomrule
\end{tabular}
\caption{Translation faithfulness analysis}
\label{table:ablation-translation-faithfulness}
\end{table}

We note that the gap is modest but directionally consistent. Better translation quality (higher mean faithfulness with lower standard deviation) is associated with more reliable verification. This observation suggests that the translation quality is directly associated with verification performance.

\section{Split between mechanical verifications and LLM-audits}
\label{app:split-mechanical-llm}
VeryTrace uses hybrid verification approaches for both the constraint verification (Sec.~\ref{sec:constraint-verification}) and COMPUTE step verification (Sec.~\ref{sec:compute-check}). Namely, both mechanical and LLM-based verifications are used for these components. To directly assess how much the mechanical verification contribute in practice, here we complement the main evaluation results from Sec.~\ref{sec:eval} by reporting the split between mechanical and LLM-based verifications we observed. In Table~\ref{table:mechanical-llm-split}, we report the split under the evaluation trials with \qwenthink{} model across all benchmarks. 

\begin{table}
\centering
\begin{tabular}{lccccc}
    \toprule
    Benchmarks & Mechanical Constraint Verification \% & Mechanical COMPUTE Verification \% \\
     \midrule
    AIME 2025 & 10.8 & 80.7 \\
    Planning-S  & 43.3 & 62.3 \\
    Planning-M  & 40.8 & 66.0 \\
    Planning-L  & 35.6 & 71.2 \\
    CLUTRR  & 0.8 & 2.8 \\
\bottomrule
\end{tabular}
\caption{Split between mechanical and LLM-based verifications across all benchmarks.}
\label{table:mechanical-llm-split}
\end{table}

We note that mechanical verification is used heavily where executable structure exists (especially AIME and BabyBench), while CLUTRR is largely semantic and therefore relies more on LLM audits. This matches the observation outlined in our manuscript where the largest gains are observed in math and planning, with a smaller margin on CLUTRR. Further, notice that given the low mechanical verification percentage in CLUTRR, VeryTrace still performed competitively against other baselines. This suggests despite heavily relying on LLM-audits on domains like CLUTRR, VeryTrace still benefits from the structure introduced by the DSL.

\section{Evaluation of VeryTrace's standalone verification capability on ProcessBench}
The verifier quality of VeryTrace directly impact the downstream performance. We further provide a standalone verifier evaluation using the ProcessBench~\cite{zheng2025processbench} under the GSM8K split. This small-scale verification is to show how well the verifier performs on a benchmark outside ones from the main evaluation as a standalone component. We report the results in Table~\ref{table:processbench}.

\begin{table}
\centering
\begin{tabular}{lccc}
    \toprule
    Metrics & Values \% \\
     \midrule
    Acceptance precision & 0.895 \\
    Acceptance recall  & 0.886 \\
    Acceptance F1  & 0.891\\
    Error-detection precision  & 0.895\\
    Error-detection recall  & 0.903\\
    Error-detection F1  & 0.899\\
    False accept rate  & 0.097\\
    False reject rate  & 0.114\\
    Overall accuracy  & 0.895\\
\bottomrule
\end{tabular}
\caption{VeryTrace's verifier performance on ProcessBench~\cite{zheng2025processbench}.}
\label{table:processbench}
\end{table}

We note that VeryTrace's verifier achieves a competitive false accept rate (9.7\%) and false reject rate (11.4\%), presenting itself as a strong verification mechanism.

\section{Limitations and Future Work}
\label{app:limitations}

\paragraph{Verification cost and scalability.}
A primary limitation of our step-level hybrid verification is runtime and resource usage. Since structural and constraint checks are performed \emph{per step}, the total number of verification calls scales linearly with the number of steps in the trace (see \ Sec.~\ref{sec:compute-check} and Sec.~\ref{sec:deduce-check}). This scaling is acceptable for moderately sized traces, but can become costly for extremely long traces or iterative feedback loops that require multiple conversion--verification rounds.

\paragraph{Dependence on semantic auditing for deductive steps.}
While most \texttt{compute} steps can be verified efficiently by deterministically executing (or re-evaluating) the extracted mathematical expressions, \texttt{deduce} steps often involve semantic transformations that are difficult to validate purely with symbolic execution. In our current pipeline, such steps may require an LLM-based audit or other semantic checker, which increases cost and can introduce stochasticity into verification outcomes.

\paragraph{Expressiveness of the inference schema library.}
Our current set of structured inference schema is intentionally small to keep compilation and verification tractable. As a consequence, certain forms of reasoning may be forced into coarse \texttt{deduce} steps (or into underspecified natural-language explanations), reducing the granularity of verifiable structure. Extending the schema library to cover richer algebraic transformations, case analysis, induction, and domain-specific semantic operators is an important direction for increasing task coverage and reducing reliance on black-box semantic auditing.

\paragraph{Future work.}
We plan to (i) improve efficiency via caching, incremental verification, and selective checking policies (e.g., prioritizing steps that affect constraints or goal satisfaction); (ii) integrate stronger formal back-ends such as SMT solvers and/or Lean-style provers to formalize and discharge a larger fraction of semantic deductions; and (iii) grow the DSL and deduction schema set while maintaining a clean compilation interface, enabling broader task coverage and finer-grained error localization.

\section{Prompt Templates}
\label{app:prompt-templates}

For reproducibility, we document the \emph{exact} prompt templates used in our experiments for the three LLM-facing stages of the VeryTrace pipeline:
(B.1) the \emph{user LLM} that produces a structured, step-decomposed reasoning trace;
(B.2) the \emph{context extractor} that reads \emph{only} the original problem statement and outputs a structured context (facts/assumptions/constraints/state/goal);
and (B.3) the \emph{DSL converter} that combines prompt-derived context with the chain-of-thought to emit a ReasonLang JSON trace.

All templates below contain placeholders (e.g., \texttt{\{problem\}}, \texttt{\{prompt\_text\}}, \texttt{\{cot\_text\}}) that are filled at runtime.

\subsection{User LLM Prompt}
\label{app:prompt-user-llm}

We standardize the user LLM output to facilitate deterministic parsing and downstream conversion. In particular, we enforce (i) explicit \emph{premises} per step, (ii) explicit \emph{assumptions used}, and (iii) explicit \emph{state updates} when variables change. We also request an \emph{atomic} decomposition into small steps to improve step-level localization during verification.

\begin{lstlisting}[style=promptstyle]
You are a careful reasoner solving complex math competition problems.

You will be given a problem from the AIME 2025 competition that requires step-by-step reasoning to solve.

Problem:
{problem}

You MUST format your reasoning according to the following rules and exact format:
- First, write down the following in the exact format:
        PROBLEM STATEMENT: ...
        GIVEN FACTS AND VARIABLES: ...
        ALL ASSUMPTIONS: ...
        CONSTRAINTS (IF ANY): ...
        UNKNOWNS: ...
- Reason step by step. Each step must be explicitly numbered and must list out the following in the exact format:
        Step <step_number>:
            PREMISES: fact1, fact2, ..., factn
            CALCULATION/REASONING: ...
            ASSUMPTIONS USED (IF ANY): ...
            UPDATE IN REASONING VARIABLES (IF ANY): ...
- **BE ATOMIC**: break down reasoning into smallest atomic steps, the more granular decomposition the better.

At the end, output the final answer on its own line in the exact format:

FINAL_ANSWER: <integer>

Note: The answer must be an integer (no decimals, no fractions). 
The FINAL_ANSWER line must be the last line of your answer.
\end{lstlisting}

\subsection{Context Extraction Prompt}
\label{app:prompt-context-extraction}

To reduce confirmation bias in the conversion stage, we first extract context from the problem statement \emph{without} giving the model access to the chain-of-thought. The extractor is instructed to not solve the problem, but instead produce a complete context bundle: initial facts, persistent assumptions, verifiable constraints (with mathematical expressions whenever possible), an initial state for tracked variables, and a precise goal.

\begin{lstlisting}[style=promptstyle]
You are a reasoning context analyzer. Your job is to extract ALL relevant context from a problem statement that would be needed for reasoning.

**CRITICAL**: You are given ONLY the problem statement. DO NOT try to solve the problem. Your goal is to extract context that would be needed for ANY reasoning approach.

**Problem Statement:**
{prompt_text}

**Your Task:**
Extract all relevant context from the problem statement above into a structured JSON format:

```json
{{
  "initial_facts": [
    {{"id": "fact_1", "content": "...", "entities": [...]}}
  ],
  "assumptions": [
    {{"id": "a_1", "content": "..."}}
  ],
  "constraints": [
    {{
      "id": "c1",
      "description": "Natural language description",
      "constraint_expr": "Mathematical expression (if applicable)",
      "scope": "invariant|goal"
    }}
  ],
  "initial_state": {{
    "variable_name": value
  }},
  "goal": "..."
}}
```

**CRITICAL DEFINITIONS - Read these carefully:**

1. **initial_facts**: Given facts that describe the INITIAL or STARTING conditions.
   - These are facts that may only hold at the beginning and can change during reasoning.
   - Examples: "Betty starts with $50", "Robot is at position (0, 0)", "John has 5 apples initially"
   - If a variable is associated or defined with a fact statement, it should also be included in the context. Example: "a: 5 is an odd number", the entire definition and statement should be included. 
   - Format: {{"id": "fact_1", "content": "...", "entities": [...]}}

2. **assumptions**: Facts assumed to be true THROUGHOUT ALL reasoning steps.
   - These are persistent truths that don't change, unlike initial_facts.
   - These are facts taken as given, NOT conditions to verify (those are constraints).
   - Look for both EXPLICIT assumptions (stated in the problem) and IMPLICIT assumptions (implied by the problem context).
   - Examples:
     * "All apples are whole apples (no fractions)" - assumption about the nature of objects
     * "The robot can only move one step at a time" - assumption about capabilities
     * "Time progresses linearly" - assumption about the environment
     * "The coordinate system has y-axis pointing north" - assumption about the reference frame
     * "Operations are performed left-to-right" - assumption about execution order
   - Format: {{"id": "a_1", "content": "..."}}
   - **BE THOROUGH**: Extract ALL assumptions, both explicit and implicit, that underpin any reasoning approach.

3. **constraints**: Conditions that must be VERIFIED or SATISFIED during reasoning.
   - These are NOT assumptions (facts taken as given), but conditions to CHECK.
   - Each constraint has:
     * "description": Natural language explanation
     * "constraint_expr": Mathematical expression MUST uses variables from initial_state (e.g., "x < 10", "position[0] != 2")
       - Leave empty if cannot be expressed mathematically
     * "scope": "invariant" (must hold at ALL steps) or "goal" (must hold only at FINAL step)
   - Examples: "The agent cannot go to position (2, 3)", "Total must not exceed 100"
   - Format: {{"id": "c1", "description": "...", "constraint_expr": "...", "scope": "invariant|goal"}}

4. **initial_state**: Initial values of variables mentioned in the problem.
   - Extract starting values for quantitative reasoning.
   - Examples: {{"position": [0, 0], "money": 50, "direction": 0}}

5. **goal**: The question or goal to be answered.

**CRITICAL INSTRUCTIONS:**
- **Extract EVERYTHING**: Be thorough - extract all facts, assumptions, and constraints that might be relevant.
- **Do NOT solve**: You are NOT solving the problem, just extracting context.
- **Look for implicit assumptions**: Don't just extract what's explicitly stated - also identify implicit assumptions the problem makes.
- **Differentiate carefully**: 
  * initial_facts = starting conditions that may change
  * assumptions = persistent truths throughout reasoning
  * constraints = conditions to verify/satisfy
- **Use mathematical expressions**: For constraints, maximize use of mathematical expressions in constraint_expr field using variables from initial_state.
- **Constraint_expr variables MUST befrom initial_state**: All constraint_expr must use variables from initial_state.
- Constraint_expr must be a boolean expression that is true if satisfied.
**Extract goal precisely**: Capture exactly what the problem is asking for.

Output ONLY the JSON structure, no other text. 
**IMPORTANT**: Your final output MUST start with ```json and end with ```.
\end{lstlisting}

\clearpage
\subsection{DSL Conversion Prompt}
\label{app:prompt-dsl-conversion}

Given the pre-extracted context and the original chain-of-thought, the converter (i) takes the \emph{union} of prompt-derived context and any additional assumptions/constraints introduced in the trace, (ii) decomposes the trace into atomic DSL steps with explicit premises, and (iii) emits a structured JSON trace suitable for subsequent structural and constraint verification.

\begin{lstlisting}[style=promptstyle]
You are a reasoning trace analyzer. Your job is to convert a natural language chain-of-thought (COT) reasoning trace into a structured JSON format.

**Original Problem/Prompt:**
{prompt_text}

**Pre-Extracted Context from Prompt:**
The following context has been extracted from the problem statement by analyzing it independently (without looking at the COT):
{extracted_context}

**Chain-of-Thought Reasoning:**
{cot_text}

**Your Task:**
Convert the reasoning trace into a structured JSON format with the following structure:

```json
{{
  "context": {{
    "initial_facts": [
      {{"id": "fact_1", "content": "...", "entities": [...]}}
    ],
    "goal": "...",
    "constraints": [
      {{
        "id": "c1", 
        "description": "Natural language description", 
        "constraint_expr": "Mathematical expression like 'x < 10' or 'position[0] != 2'",
        "scope": "invariant|goal"
      }}
    ],
    "assumptions": [
      {{"id": "a_1", "content": "..."}}
    ],
    "initial_state": {{
      "variable_name": value,
      "another_variable": [x, y]
    }}
  }},
  "steps": [
    {{
      "step_number": 1,
      "inference_type": "assume|deduce|compute|conclude",
      "claim": "...",
      "premises": ["fact_1", "step_2"],
      "reasoning": "...",
      "assumptions": ["...", ...],
      "updates": {{"variable_name": new_value, ...}},
      "compute_expr": "expression for compute steps",
      "deduction_rule": "modus_ponens|direct|transitivity|...",
      "deduction_args": {{"arg1": "...", "arg2": "..."}} for deduce steps,
      "is_final": false
    }}
  ],
  "conclusion": {{
    "content": "..."
  }}
}}
```

**CRITICAL INSTRUCTIONS FOR COMBINING CONTEXT:**

The pre-extracted context above was created by analyzing the prompt ONLY (without seeing the COT). The COT may also mention or use some facts, assumptions, or constraints.

Your job is to **take the UNION** of:
1. Context extracted from the prompt (provided above)
2. Any additional context mentioned or used in the COT

**How to combine:**
- **initial_facts**: Use the ones from pre-extracted context. The COT typically doesn't add new initial facts.
- **assumptions**: COMBINE (union) assumptions from pre-extracted context AND any additional assumptions mentioned in the COT.
  - If the COT explicitly states an assumption not in the pre-extracted context, ADD it.
  - If an assumption appears in both, keep it once (no duplicates).
  - Preserve IDs from pre-extracted context; assign new IDs (a_N+1, a_N+2, ...) for new assumptions from COT.
- **constraints**: COMBINE (union) constraints from pre-extracted context AND any constraints mentioned in the COT.
  - Same rules as assumptions: add new ones, avoid duplicates.
  - Preserve IDs from pre-extracted context; assign new IDs for new constraints.
- **initial_state**: Use from pre-extracted context, or enhance if COT provides additional variable values.
- **goal**: Use from pre-extracted context.

**Instructions:**
1. Use initial_facts from pre-extracted context with IDs like "fact_1", "fact_2", etc.

2. **Use initial_state from pre-extracted context**: The initial state variables have been extracted from the problem statement.

3. Use goal from pre-extracted context.

4. **Decompose COT into atomic steps**:
   - Each step should be a single inference, calculation, or deduction.
   - If the original COT combines multiple reasoning operations in one step, SPLIT them into separate steps.
   - Examples of what to split:
     * "Calculate X and then use it to determine Y" --> Split into: Step 1 (calculate X), Step 2 (determine Y using X)
     * "From A and B, we conclude C and D" --> Split into: Step 1 (from A and B conclude C), Step 2 (from C conclude D)
     * Multiple calculations in sequence --> Each calculation is a separate step

5. **COMBINE constraints** from pre-extracted context and COT:
   - The pre-extracted context already has constraints with mathematical expressions where possible
   - If the COT mentions additional constraints not in pre-extracted context, ADD them
   - For new constraints:
     * **MAXIMALLY USE MATHEMATICAL EXPRESSIONS**: This is the MOST IMPORTANT requirement
     * For EACH constraint, try to express it as a mathematical expression using:
       - Comparison operators: <, >, ==, !=, <=, >=
       - Logical operators: and, or, not
       - Arithmetic: +, -, *, /
       - Array/tuple indexing: position[0], position[1]
       - Variable names that will be tracked in reasoning state
     * **constraint_expr field**: mathematical expression MUST uses variables from initial_state
     * **description field**: Natural language explanation of the constraint
     * ONLY use natural language (leave constraint_expr empty) if the constraint CANNOT be expressed mathematically
   - **Constraint structure**:
     * Give each constraint a unique ID: use IDs from pre-extracted context, assign new IDs like "c_N+1" for additions
     * Give each constraint a scope: "invariant" or "goal"
       - "invariant": Must be satisfied at ALL reasoning steps
       - "goal": Must be satisfied only at the FINAL reasoning step (success criteria)
     * **BE SPECIFIC**: Use actual values, not placeholders (e.g., "position[0] != 2" not "position[0] != x")
     * **BE ATOMIC**: Each constraint must be a single condition
     * **USE VARIABLES FROM THE REASONING STATE**: variables in constraint_expr MUST be from the initial_state

6. **COMBINE assumptions** from pre-extracted context and COT:
   - The pre-extracted context already has assumptions
   - If the COT explicitly mentions or relies on additional assumptions not in pre-extracted context, ADD them
   - Format: {{"id": "a_N+1", "content": "..."}}
   - Preserve IDs from pre-extracted context; assign new sequential IDs for additions

7. For each reasoning step:
   - **inference_type**: Classify the step as one of:
     * "assume" - Establishing a fact from context/premises
     * "deduce" - Logical deduction or inference (MUST provide deduction_rule and deduction_args) from premises
     * "compute" - Performing calculations (MUST provide compute_expr and updates)
     * "conclude" - Final conclusion
   
   - **claim**: The main assertion being made in this step
   
   - **premises**: CRITICAL - List the IDs of facts/steps this depends on using ONLY these formats:
     * "fact_1", "fact_2", etc. for initial facts from context
     * "step_1", "step_2", etc. for previous reasoning steps
     * Example: ["fact_1", "step_2"] means this step uses fact_1 and the result from step_2
     * DO NOT write natural language descriptions - ONLY use "fact_#" or "step_#" format
   
   - **reasoning**: Natural language explanation of why this step follows
   
   - **assumptions**: Any assumptions made (can reference assumption IDs from context)
   
   - **FOR COMPUTE STEPS ONLY - BOTH REQUIRED**:
     * **compute_expr**: An ASSIGNMENT expression in the form "variable = expression"
       - MUST contain an equals sign (assignment operator)
       - The RHS expression must use ONLY variables from the current reasoning state
       - Examples:
         * "position = [position[0] + 1, position[1]]" (updates entire variable)
         * "x = x + 1" (updates existing variable)
         * "new_x = position[0] + 1" (creates new variable)
       - INVALID: "position[0] + 1" (missing assignment)
       - INVALID: "x + 1" (missing assignment)
     * **updates**: The resulting variable changes after the assignment
       - Examples: {{"position": [4, 3]}}, {{"x": 2}}, {{"new_x": 4}}
       - Must introduce new variables OR update existing ones
       - The updates must reflect the state changes from the assignment
   
   - **FOR DEDUCE STEPS ONLY**:
     * MUST provide at least one premise.
     * **deduction_rule**: The logical rule being applied, one of:
       - "modus_ponens" - A -> B, A => B
       - "conjunction" - A, B => A and B
       - "direct" - Conclusion follows directly from premises
       - "transitivity" - A = B, B = C => A = C
     * **deduction_args**: Structured arguments for the deduction rule
       - For "modus_ponens": {{"conditional": "if A then B", "antecedent": "A", "consequent": "B"}}
       - For "direct": {{"premise_content": "...", "conclusion": "..."}}
       - For "transitivity": {{"first_equality": "A = B", "second_equality": "B = C", "conclusion": "A = C"}}
       - Adapt format based on the deduction rule
     * **updates**: MUST provide updates that introduce new variables or modify existing ones

   
   - **is_final**: true only for the last step

8. Extract the final conclusion
   - **content**: The content must have enough information to fully answer the original problem.
   - **BE THOROUGH**: The content must be thorough and complete. Avoid a single-word answer.

**CRITICAL RULES FOR PREMISES:**
- ALWAYS use "step_#" to reference previous steps (e.g., "step_1", "step_2")
- ALWAYS use "fact_#" to reference initial facts (e.g., "fact_1", "fact_2")
- NEVER write natural language in premises like "Initial state is X" or "Betty has $50"
- Each premise MUST be exactly "step_#" or "fact_#"
- Deduce steps must have at least one premise.

**CRITICAL RULES FOR INFERENCE_TYPE:**
- Use "compute" ONLY when the step directly evaluates a mathematical expression:
  * A "compute" step MUST provide BOTH "compute_expr" (executable expression) AND "updates" (results)
  * The compute_expr must use ONLY variables from the reasoning state
  * The updates must introduce new variables OR modify existing ones
  * Examples: "position[0] + 1", "total - 15", "[x, y + 1]"
  
- Use "deduce" for logical reasoning, including reasoning about transformations:
  * A "deduce" step MUST provide BOTH "deduction_rule" AND "deduction_args"
  * A "deduce" step SHOULD provide "updates" when it modifies reasoning state
  * Updates are optional for intermediate logical conclusions
  * Use "deduce" when the step involves understanding semantics or rules
  
- Use "assume" for establishing facts from the problem statement
- Use "conclude" for the final answer

**KEY DISTINCTION:**
- "compute": Direct calculation with existing values --> "position[0] + 1" 
- "deduce": Reasoning about what transformation applies --> "turning left from east means facing north, so new direction is 3"
- MAXIMALLY use "compute" steps when math is involved.

**CRITICAL RULES FOR COMPUTE STEPS:**
- MUST include "compute_expr": an ASSIGNMENT expression with format "variable = expression"
- MUST include "updates": the resulting variable changes
- The compute_expr MUST contain an equals sign (=) for assignment
- The RHS of the assignment must reference ONLY existing state variables
- Examples of VALID expressions:
  * "x = x + 1" (if x exists in state)
  * "position = [position[0] + 1, position[1]]" (if position exists)
  * "new_total = old_total + amount" (if old_total and amount exist)
- Examples of INVALID expressions:
  * "position[0] + 1" (missing assignment - which variable does this update?)
  * "y = z + 1" (if z doesn't exist in state)

**CRITICAL RULES FOR DEDUCE STEPS:**
- MUST include "deduction_rule": one of the defined logical rules
- MUST include "deduction_args": structured arguments for that rule
- SHOULD include "updates": state changes from the deduction (optional for intermediate steps)
- The deduction_args must clearly specify the logical structure
- Use "direct" rule for straightforward logical conclusions
- Examples:
  * For orientation changes: {{"deduction_rule": "direct", "deduction_args": {{"premise_content": "agent faces east (direction=0), turning left rotates counter-clockwise", "conclusion": "agent now faces north (direction=3)"}}, "updates": {{"direction": 3}}}}
  * For movement: {{"deduction_rule": "direct", "deduction_args": {{"premise_content": "agent at (3,4) facing west, moving forward goes west", "conclusion": "agent moves to (2,4)"}}, "updates": {{"position": [2,4]}}}}
  * For intermediate logic: {{"deduction_rule": "direct", "deduction_args": {{"premise_content": "target is east of agent", "conclusion": "agent needs to turn to face east"}}, "updates": {{}}}}
  * For modus_ponens: {{"deduction_rule": "modus_ponens", "deduction_args": {{"conditional": "if agent faces east and turns left then agent faces north", "antecedent": "agent faces east and turns left", "consequent": "agent faces north"}}, "updates": {{"direction": 3}}}}

**IMPORTANT:**
- **COMBINE context**: Take union of pre-extracted context and any additional context from COT
- **Avoid duplicates**: If the same assumption/constraint appears in both, keep it once
- **Preserve IDs**: Keep IDs from pre-extracted context; assign new sequential IDs for additions
- **Extract constraints with ACTUAL values**: Use specific values from the problem  
- **No Magic Constraint Variables**: Do not use variables that are not in the initial_state in the constraint_expr
- **Track state through updates**: Every variable introduction/change should be in "updates"
- **Pay special attention** to constraints in the original prompt that the COT might have ignored
- Classify constraint scope correctly (invariant vs goal)
- In the "updates" field, use JSON arrays [x, y] not tuples (x, y)
- For nested structures, use arrays: [[1, 2], 3] not ((1, 2), 3)

Final Note: Output ONLY the JSON structure, no other text.
\end{lstlisting}

\subsection{Translation faithfulness evaluation prompt template}
\label{app:translation-faithfulness-prompt}

Given the natural language CoT and the translated formal DSL, we use the following prompt template to query the LLM to evaluate the faithfulness of the translation based on a score from 1 to 10.

\begin{lstlisting}[style=promptstyle]
You are an expert evaluator assessing the faithfulness of a formal DSL (Domain-Specific Language) translation of a natural-language chain-of-thought (COT) reasoning trace.

Important: The DSL has already passed syntax validation. it is structurally well-formed. Your task is NOT to check for syntax errors. Instead, you must evaluate whether the DSL is a **faithful semantic translation** of the original NL COT, regardless of whether the original NL COT itself is logically correct.

**Brief DSL schema overview (for reference only):**
The DSL is a JSON object with these key sections:
- **"initial_facts"**: A list of facts extracted from the problem statement / NL COT preamble. Each has an "id" and "content".
- **"assumptions"**: A list of assumptions the reasoner stated or relied on. Each has an "id" and "content".
- **"constraints"**: A list of constraints that the reasoning should respect. Each has a "description" and optional "constraint_expr".
- **"initial_state"**: The starting state of the reasoning (e.g., position, direction).
- **"goal"**: The goal the reasoner is trying to achieve.
- **"steps"**: A list of reasoning steps. Each step has:
    - "step_number": ordinal index
    - "inference_type": one of "compute", "deduction", "observation", etc.
    - "claim": what the step asserts
    - "premises": references to prior steps or facts used
    - "reasoning": the reasoning or calculation performed
    - "assumptions": assumptions referenced
    - "compute_expr" (if applicable): a formalized expression of the computation
    - "updates": state changes produced by this step
- **"conclusion"**: The final conclusion of the reasoning.

**Original Problem / Prompt:**
{prompt_text}

**Original Natural-Language COT:**
{nl_cot}

**Translated DSL:**
{dsl_json_str}

**Your Evaluation Task:**
Assess whether the DSL is a faithful translation of the original NL COT. "Faithful" means the DSL accurately captures what the NL COT says - not whether the NL COT is correct. Specifically:

**Faithfulness Score (1-10):** How well does the DSL capture the content and reasoning of the original NL COT?
1-2: The DSL bears little resemblance to the NL COT.
3-4: Major portions of the NL COT are missing, distorted, or fabricated in the DSL.
5-6: The DSL captures the general approach but has notable omissions or additions.
7-8: The DSL is mostly faithful with minor omissions, additions, or rephrasings.
9-10: The DSL is a near-perfect or perfect translation of the NL COT.

Respond in EXACTLY this format (no other text):
FAITHFULNESS_SCORE: <integer 1-10>
\end{lstlisting}


\end{document}